\documentclass[runningheads]{llncs}

\newif\ifdraft\drafttrue
\newif\ifinlineref\inlinereffalse
\newif\iffinal\finalfalse
\newif\ifextended\extendedfalse
\newif\ifdotikz\dotikzfalse

\def\uminus{\setbox0=\hbox{$\cup$}\rlap{\hbox
    to\wd0{\hss\raise0.3ex\hbox{$\scriptscriptstyle{-}$}\hss}}\box0}

\def\capminus{\setbox0=\hbox{$\cap$}\rlap{\hbox
    to\wd0{\hss\raise0.3ex\hbox{$\scriptscriptstyle{-}$}\hss}}\box0}

\newcommand{\naf}{\mathbf{\mathop{not}\,}}

\usepackage{url}
\usepackage{color}
\usepackage{amssymb}

\definecolor{darkgreen}{rgb}{0,0.7,0}
\definecolor{darkblue}{rgb}{0,0,0.7}
\definecolor{darkred}{rgb}{0.7,0,0}

\usepackage{comment}
\long\def\comment#1{}

\usepackage{amsmath}
\usepackage[ruled,vlined]{algorithm2e}
\usepackage{paralist}
\usepackage{multirow}
\usepackage{booktabs}

\newcounter{myenumctr}

\begin{document}

\title{Constraint Monotonicity, Epistemic Splitting \\ and Foundedness
  Can in General Be Too Strong \\  in Answer Set Programming\thanks{This paper
  is presented at TAASP 2020: Workshop on Trends and Applications of Answer Set Programming.}}

\titlerunning{Constraint Monotonicity, Epistemic Splitting and Foundedness Are Too Strong}

\author{Yi-Dong Shen\inst{1} \and
Thomas Eiter\inst{2}}
\authorrunning{Y. D. Shen and T. Eiter}
%
\institute{State Key Laboratory of Computer Science, Institute of
Software, Chinese Academy of Sciences,  
Beijing 100190, China (\email{ydshen@ios.ac.cn}) \and
Institute of Logic and Computation, Vienna University of Technology
(TU Wien), \\ Favoritenstra{\ss}e 9-11, A-1040 Vienna, Austria 
(\email{eiter@kr.tuwien.ac.at})}

\maketitle  

\begin{abstract}
Recently, the notions of {\em subjective constraint monotonicity}, 
{\em epistemic splitting}, and {\em foundedness}\/
have been introduced
for epistemic logic programs, with the aim
to use them as main criteria 
respectively intuitions to compare
different answer set semantics 
proposed in the literature on how they comply with these intuitions.
In this note, we consider these three notions and
demonstrate on some examples that 
they may be too strong
in general
and may exclude
some desired answer sets respectively
world views.
In conclusion, these
properties should not be regarded
as mandatory properties that every answer set semantics
must satisfy in general.  
\end{abstract}

\section{Introduction}
\label{int}
In a seminal paper, 
Gelfond \cite{Gelfond91} introduced the notion of {\em epistemic specifications}
which are disjunctive logic programs extended 
with two epistemic modal operators ${\bf K}$ and ${\bf M}$.
Informally, for a formula $F$ and a collection $\cal A$ of interpretations,
${\bf K} F$ is true in $\cal A$ if
$F$ is true in every $I\in {\cal A}$, and 
${\bf M} F$ is true in $\cal A$ if
$F$ is true in some $I\in {\cal A}$. 
An epistemic specification/program $\Pi$
consists of rules of the form
\begin{equation}
\label{equa-1}
L_1\, \mid \cdots \mid \, L_m \leftarrow G_1 \wedge \cdots \wedge G_n
\end{equation}
where each $L$ is an {\em object literal} that is either
an atom $A$ or its strong negation $\sim$$A$,
and
each $G$ is an object literal, 
a default negated literal of the form $\neg L$,%
\footnote{We use here $\neg$ for weak negation (alias default
negation), as in early papers on logic programming.}
or a {\em modal literal} of the form ${\bf K} L$,
$\neg {\bf K} L$,  ${\bf M} L$ or  
$\neg {\bf M} L$.
A rule (\ref{equa-1}) is called a {\em constraint} if its head is $\bot$,
and called a {\em subjective constraint} if additionally
each $G$ is a modal literal.
$\Pi$ is a {\em non-epistemic program} (or an {\em
answer-set program}) if it contains no modal literals.

Gelfond defined then the first answer set semantics for 
an epistemic program $\Pi$ as follows \cite{Gelfond91}.
Given a collection $\cal A$ of interpretations
as an assumption, $\Pi$ is transformed
into a {\em modal reduct} $\Pi^{\cal A}$ 
w.r.t.\ $\cal A$ by 
first removing all rules with a modal literal $G$ 
that is not true in $\cal A$, 
then removing the remaining modal literals.
The assumption 
$\cal A$ is defined to be a {\em world view} of $\Pi$ 
if it coincides with the collection of answer sets of $\Pi^{\cal A}$ 
under the {\em GL-semantics} defined in \cite{GL91}. 

It turned out that the above semantics
for epistemic programs has both 
{\em the problem of unintended world views with recursion through} {\bf K}
and {\em the problem due to recursion through} {\bf M} \cite{Gelfond2011,Kahl14}.
For the first problem, an illustrative example
is $\Pi = \{p\leftarrow {\bf K} p\}$;
under the above semantics
$\Pi$ has two world views ${\cal A}_1 = \{\emptyset\}$
and ${\cal A}_2 = \{\{p\}\}$, where 
as commented in \cite{Gelfond2011},  
${\cal A}_2$ is undesired.
For the second problem, a typical example
is $\Pi = \{p\leftarrow {\bf M} p\}$;
by the above semantics
$\Pi$ has two world views ${\cal A}_1 = \{\emptyset\}$
and ${\cal A}_2 = \{\{p\}\}$, where
as commented in \cite{Kahl14}, ${\cal A}_1$ may be undesired.

To address the two problems, several approaches 
have been proposed \cite{Kahl14,KahlGelfond2015,Fa2015,ShenEiter16}.
In particular, Shen and Eiter\cite{ShenEiter16} presented an approach 
that significantly differs from the others in the following three
aspects.

\begin{enumerate}[(i)]
\item They introduced the modal operator $\naf$ to directly express 
epistemic negation, where $\naf F$ expresses that there is no
evidence proving that $F$ is true. 
Modal formulas ${\bf K} F$ and ${\bf M} F$
are viewed as shorthands for $\neg \naf F$ and $\naf \neg F$,
respectively.

\item 
Due to having the modal operator $\naf$ to express 
epistemic negation, they further proposed to apply 
epistemic negation to minimize the knowledge in
world views, a novel principle they named
{\em knowledge minimization
with epistemic negation}.
It is based on the principle of knowledge minimization
with epistemic negation that they presented a 
completely new definition of 
world views, which
are free of both the problem with recursion through {\bf K} and 
the problem through {\bf M}.

\item 
Their approach is {\em generic} in the sense that it
can be used to extend any of the
existing answer set semantics 
for non-epistemic programs, such as those defined in
\cite{Pearce06,PDB07,Truszczynski-aij10,FaberPL11,FerrarisLL11,ShenWEFRKD14,ShenE19}, 
but also novel ones so they may be extended to an
answer set semantics for epistemic programs.
\end{enumerate}

Very recently,  
some researchers \cite{iclp-KahlL18,lpnmr-CabalarFC19,CabalarFC19a}
introduced the notions of {\em subjective constraint monotonicity}, 
{\em epistemic splitting}, and {\em foundedness}
for epistemic programs, aiming to
use them as main criteria/intuitions to compare
different answer set semantics proposed in the literature on how they comply with these intuitions.
Specifically, 
they criticized
the semantics defined in 
\cite{Kahl14,KahlGelfond2015,Fa2015,ShenEiter16}, saying that
these semantics do not satisfy the three properties.

In this note, we clarify the matter by demonstrating on some
example programs that 
these three properties 
may be too strong and may exclude
some desired answer sets/world views.
Our conclusion is that for this reason these
properties should not be
used as mandatory properties that every
answer set semantics must satisfy
in general. 

For the remainder of this note, we assume that the reader is
familiar with non-monotonic logic programs in general and with answer
set semantics for such programs in particular. We refrain here
from providing formal definitions of answer sets and of world views of
epistemic logic programs; for our concerns, it is sufficient to 
assume that the programs are formulated over a set $V$ of propositional
atoms together with the special atoms $\top$ (truth) and $\bot$ (falsity).
An answer set of an answer-set program $\Pi$ is an
interpretation $I\subseteq V$ that satisfies respective conditions,
where the standard definition is {\em GL-semantics} \cite{GL91}. 
Similarly, a world view is a non-empty collection $\mathcal{A}
\subseteq 2^V$ of interpretations that must satisfy respective
conditions such as those in \cite{Gelfond91}, which yield the 
G91-semantics for epistemic logic programs. Numerous further proposals
for semantics have been made, cf.\ \cite{Gelfond2011,Truszczynski11,Kahl14,KahlGelfond2015,Fa2015,ShenEiter16,iclp-KahlL18,lpnmr-CabalarFC19,CabalarFC19a,jelia-Su19,DBLP:journals/corr/abs-1909-08233,DBLP:journals/ai/SuCH20}.

\section{Subjective constraint monotonicity is too strong, while the requirement of epistemic splitting is even more restrictive}

A semantics for epistemic logic programs
is said to satisfy {\em subjective constraint monotonicity}
if for any epistemic program $\Pi$ and subjective constraint $C$,
a world view of $\Pi\cup \{C\}$ is also a world view of $\Pi$;
in other words, adding any constraint $C$ to $\Pi$
would never introduce new world views.
The {\em epistemic splitting} property is even more restrictive in the sense 
that every semantics satisfying epistemic splitting also satisfies 
subjective constraint monotonicity, as has been shown in \cite{lpnmr-CabalarFC19}.

As a typical example, let $\Pi= \{p\mid q\}$, which has a unique world view
$\{\{p\},\{q\}\}$. Then subjective constraint monotonicity
requires that for any subjective constraint $C$,
$\Pi\cup \{C\}$ should either have a unique world view $\{\{p\},\{q\}\}$
or have no world view. 
For example, under subjective constraint monotonicity
the following program
\begin{align}
 \Pi_1:\quad & p\mid q  \tag{$r_1$}\\
             & \bot \leftarrow \neg {\bf K} p  \tag{$C$}
\end{align}
has no world view,
as the only world view $\{\{p\},\{q\}\}$ of $\Pi= \{p\mid q\}$ 
is not a model of $\Pi_1$. 
Note that under the semantics of  
\cite{Kahl14,KahlGelfond2015,Fa2015,ShenEiter16},
$\Pi_1$ has a world view ${\cal A} = \{\{p\}\}$. 
It is argued in \cite{iclp-KahlL18,lpnmr-CabalarFC19,jelia-Su19}
that $\{\{p\}\}$ should not be a world view of $\Pi_1$
because it violates subjective constraint monotonicity. 

We comment that the requirement of constraint monotonicity (resp. epistemic splitting),
i.e., adding constraints to a logic program
should not introduce new answer sets/world views,
may be
too strong in general and may exclude
some desired answer sets/world views,
as demonstrated in the following examples.

\begin{enumerate}
\item
For a non-epistemic program $\Pi$, the GL-semantics 
\cite{GL91} satisfies the constraint monotonicity property that 
adding a constraint $\bot\leftarrow body(r)$ to $\Pi$
may rule out some answer sets of $\Pi$, but would never
introduce new answer sets \cite{LTT99}. 
However, very recent research \cite{ShenE19}
reveals that the GL-semantics
may miss some desired answer sets
that violate constraint monotonicity (see Section 4.1 in \cite{ShenE19}).
As an example, consider the following
non-epistemic program:

\begin{align} 
\Pi_2:\quad & a\mid b  \tag{$r_1$} \\
            & a\leftarrow b  \tag{$r_2$}\\
& \bot \leftarrow \neg b   \tag{$C$}
\end{align}
where $C$ is a constraint.
Intuitively, the rule $r_1$ presents two alternatives 
for answer set construction, namely $a$ or $b$,
and the rule $r_2$ infers $a$ if $b$ has already been derived.
We distinguish between the following two cases.

$\hspace{.2in}$ First, suppose that we choose $a$ from $r_1$.
As $b$ is not inferred from $r_1$,
the rule $r_2$ is not applicable;
so rules $r_1$ and $r_2$ together infer 
a possible answer set $I_1=\{a\}$.
As $I_1$ does not satisfy the constraint $C$,
it is not a candidate answer set for $\Pi_2$.

$\hspace{.2in}$  Alternatively, 
suppose that we choose $b$ 
from $r_1$; then by $r_2$ we obtain
a possible answer set $I_2=\{a,b\}$. 
$I_2$ satisfies the constraint $C$, 
so it is a candidate answer set for $\Pi_2$.

$\hspace{.2in}$  As $I_2=\{a,b\}$ is the only model of $\Pi_2$,  
it is the only candidate answer set and
thus we expect $I_2$ to be an answer set of $\Pi_2$.
However, as $\Pi_2\setminus \{C\}$ has only one answer set $\{a\}$,
this desired answer set $I_2$ for $\Pi_2$
violates the constraint monotonicity property. 

\item
For epistemic programs, the requirement of subjective constraint monotonicity
(resp. epistemic splitting) may also exclude
some world views that are reasonably acceptable. As an example, consider 
the above program $\Pi_1$ again. 
As the rule $r_1=p\mid q$ offers two alternatives for answer set construction,
namely $p$ or $q$,
we can generate from 
$r_1$ two possible answer sets: $\{p\}$ and $\{q\}$.
Then we can construct from the two possible
answer sets three possible world views:
${\cal A}_1 = \{\{p\}\}$, ${\cal A}_2 = \{\{q\}\}$ and ${\cal A}_3 = \{\{p\},\{q\}\}$.
As ${\cal A}_2$ and ${\cal A}_3$ do not satisfy the constraint 
$\bot \leftarrow \neg {\bf K} p$, ${\cal A}_1$ is the only candidate world view
and thus we expect it to be a world view of $\Pi_1$.
However, this desired world view will be excluded if we enforce 
subjective constraint monotonicity.

\item
The above defined constraint monotonicity, which requires
world views of $\Pi\cup \{C\}$ to be world views of $\Pi$
satisfying $C$, amounts in essence to {\em interpreting the
constraint $C$ as a query} in the tradition of logic
programming; that is, in order to answer a goal query $Q$ against a logic
program $P$, we add the clause $\bot \leftarrow Q$ and then seek 
to derive $Q$. In  the context of epistemic logic programs, where
multiple world views are possible in general, we may view this as follows.
Let $S$ be the collection of world views of $\Pi$. A query $C$ to $\Pi$ is to find in $S$ all world views that satisfy $C$. Note that query $C$ is not involved
in the computation of any world view. This essentially differs from
adding a constraint $C$ to $\Pi$, which aims to
play a governing role in building
the collection of world views of $\Pi\cup \{C\}$; 
due to that $C$ is directly involved in the computation of every world view,
a world view of $\Pi\cup \{C\}$ is not necessarily a world view of $\Pi$.
\end{enumerate}

\section{The foundedness requirement is also too strong}

The foundedness property is defined in \cite{CabalarFC19a},
where a proposal for generalizing
the notion of {\em foundedness}\/ introduced in \cite{LeoneRS97} 
for non-epistemic programs to epistemic programs
has been made.
The GL-semantics \cite{GL91} for non-epistemic programs
also has the foundedness property.
We use examples to demonstrate that the foundedness 
requirement is too strong and may exclude some desired answer sets/world views.
For simplicity, we do not reproduce the definition of foundedness here;
the reader is referred to \cite{CabalarFC19a}.

\begin{enumerate}
\item
 Consider again the non-epistemic program $\Pi_2$ from above.
Note that for the construction of an answer set, the rule $r_1$
provides two alternatives, $a$ or $b$,
for us to choose. Let $b$ be selected from $r_1$.
Then once $b$ is established in $r_1$, 
$a$ is well-supported and thus derived from 
$r_2$. This leads to a possible answer set $I=\{a,b\}$.
As $I$ satisfies the constraint $C$,
it is a candidate answer set for $\Pi_2$.
As $I$ is the only model of $\Pi_2$,
it is the only candidate answer set for $\Pi_2$
and thus is a desired answer set of $\Pi_2$.
However, 
this desired answer set
violates the foundedness property. (It is easy to check that $\langle \{b\}, I\rangle$
is an unfounded set.)

\item
Consider the following epistemic program:
\begin{align} 
  \Pi_3:\quad & p\mid q \tag{$r_1$}\\
 &  p\leftarrow {\bf K} q \tag{$r_2$}\\
 &  q\leftarrow {\bf K} p  \tag{$r_3$}\\
 & \bot \leftarrow \neg {\bf K} p \tag{$C$}
\end{align}
As $p\mid q$ offers two alternatives for answer set construction,
namely $p$ or $q$,
we can generate from 
$r_1$ two possible answer sets: $\{p,\cdots\}$ and $\{q,\cdots\}$,
where ``$\cdots$'' stands for possible atoms 
that would be derived from the rules $r_2$ and $r_3$. 
Then we can construct
from the two possible answer sets three possible world views:
${\cal A}_1 = \{\{p,\cdots\}\}$, ${\cal A}_2 = \{\{q,\cdots\}\}$, and 
${\cal A}_3 = \{\{p,\cdots\},\{q,\cdots\}\}= \{\{p\},\{q\}\}$. Note that
the two answer sets in ${\cal A}_3$ must be different and no one is 
a proper subset of the other. 
We distinguish among the following three cases.

$\hspace{.2in}$ First, suppose that we choose
${\cal A}_1 = \{\{p,\cdots\}\}$.
Note that $p$ in ${\cal A}_1$ is established in $r_1$.
Then, as ${\cal A}_1$ satisfies ${\bf K} p$,
$q$ is well-supported in $r_3$ and thus ${\cal A}_1 = \{\{p,q\}\}$.
${\cal A}_1$ also satisfies $r_2$ and $C$,
so it is a candidate world view for $\Pi_3$.

$\hspace{.2in}$  Second, 
suppose that we choose
${\cal A}_2 = \{\{q,\cdots\}\}$.
Note that $q$ in ${\cal A}_2$ is established in $r_1$.
Then, as ${\cal A}_2$ satisfies ${\bf K} q$,
$p$ is well-supported in $r_2$ and thus ${\cal A}_2 = \{\{p,q\}\}$.
${\cal A}_2$ satisfies $r_3$ and $C$,
so it is further shown that $\{\{p,q\}\}$
is a candidate world view for $\Pi_3$.

$\hspace{.2in}$  Finally, 
suppose that we choose
${\cal A}_3 = \{\{p\},\{q\}\}$.
${\cal A}_3$ does not satisfy $C$,
so it is not a candidate world view for $\Pi_3$.

$\hspace{.2in}$  Consequently, $\{\{p,q\}\}$ is the only 
candidate world view for $\Pi_3$,
so we may expect it to be a world view of $\Pi_3$.
However, 
this desired world view
violates the foundedness property. 
(It is easy to check that $[\langle \{p\}, \{p,q\}\rangle, \langle \{q\}, \{p,q\}\rangle]$
is an unfounded set.) 
\end{enumerate}

\section{Conclusions}
The above examples demonstrate that the properties of 
subjective constraint monotonicity, epistemic splitting
and foundedness
are too strong and may exclude some desired answer sets/world views.
It was specifically
emphasized in \cite{Gelfond2011,Kahl14,ShenEiter16}
that the focus of research on answer set semantics 
for epistemic programs is how to handle the two basic problems:
\begin{enumerate}
\item The problem of unintended world views 
caused by recursion through {\bf K};
\item
The problem  of unintended world views caused
due to recursion through {\bf M}.
\end{enumerate}
 In fact, by introducing the epistemic negation operator $\naf$ and
applying the principle of knowledge minimization with epistemic
negation, Shen and Eiter \cite{ShenEiter16} has presented a principled
way to handle the two problems.  For example, the desired answer sets
respectively world views of the above programs
$\Pi_1-\Pi_3$ can all be obtained by applying the general semantics
of Definition~8 in \cite{ShenEiter16}, where the
base answer set semantics ${\cal X}$ for a non-epistemic program is
the one according to Definition~10
in \cite{ShenE19}. 
This is not to say, however,
that the Shen-Eiter approach in \cite{ShenEiter16} is superior to
the others. Similar as for the extension of answer-set
  program with aggregates (cf.\ \cite{DBLP:journals/ki/AlvianoF18} for a brief survey), there
  is a spectrum of possibilities with a range of properties and
  features. We believe that like for that extension, understanding
  the landscape of diverse approaches for answer set semantics of epistemic logic
  programs is a valuable goal, and that properties of universal
  validity may need comprehensive examinations.
  
\section*{Acknowledgments}
We thank the anonymous 
reviewers from the TAASP 2020 workshop for their helpful comments.
This work has been supported in part by NSFC grants 61976205 and 61379043, the Austrian
Science Fund (FWF) grant W1255, and the EU grant HumanE-AI-Net (ICT-48-2020-RIA 952026).

\bibliographystyle{splncs04}

\end{document}